\documentclass[letterpaper, 10 pt, conference]{ieeeconf}
\IEEEoverridecommandlockouts
\usepackage{xcolor}
\usepackage{booktabs}
\usepackage{epsfig}
\usepackage{graphicx}
\usepackage{amsmath}
\usepackage{amssymb}
\usepackage{subcaption}
\usepackage{pifont}

\definecolor{cvprblue}{rgb}{0.21,0.49,0.74}
\usepackage[breaklinks,colorlinks = true,citecolor=cvprblue, urlcolor=cvprblue]{hyperref}

\usepackage{dsfont}
\usepackage{bbm}
\usepackage{algorithm}
\usepackage{algpseudocode}

\usepackage{pifont}
\usepackage{bm}

\newcommand{\acronym}{WildLMa}

\usepackage
[sortcites,
backend=bibtex,
bibstyle=ieee,
citestyle=numeric,
sorting = nyt,
mincitenames=1,
maxcitenames=2,
natbib=true,
doi=false,
isbn=false,
url=false,
eprint=false]{biblatex}

\makeatletter

\usepackage[margin=0.75in]{geometry}

\title{\LARGE \bf WildLMa: Long Horizon Loco-Manipulation in the Wild}

\author{
    Ri-Zhao Qiu$^{*1}$, Yuchen Song$^{*1}$, Xuanbin Peng$^{*1}$, Sai Aneesh Suryadevara$^{1}$, Ge Yang$^{2}$, Minghuan Liu$^{1}$
    \\   Mazeyu Ji$^{1}$, Chengzhe Jia$^{1}$, Ruihan Yang$^{1}$, Xueyan Zou$^{1}$, Xiaolong Wang$^1$ 
    \\  $^{*}$equal contribution
    \\  $^{1}$UC San Diego $^{2}$MIT 
    \\ {\color{es-blue}{\texttt{\url{https://wildlma.github.io}}}}
}

\definecolor{es-blue}{rgb}{0,0.4,0.8}

\begin{document}

\twocolumn[{%
\renewcommand\twocolumn[1][]{#1}%
\maketitle

\vspace{-3mm}

\begin{center}
    \centering
    \captionsetup{type=figure}
    \includegraphics[width=\linewidth]{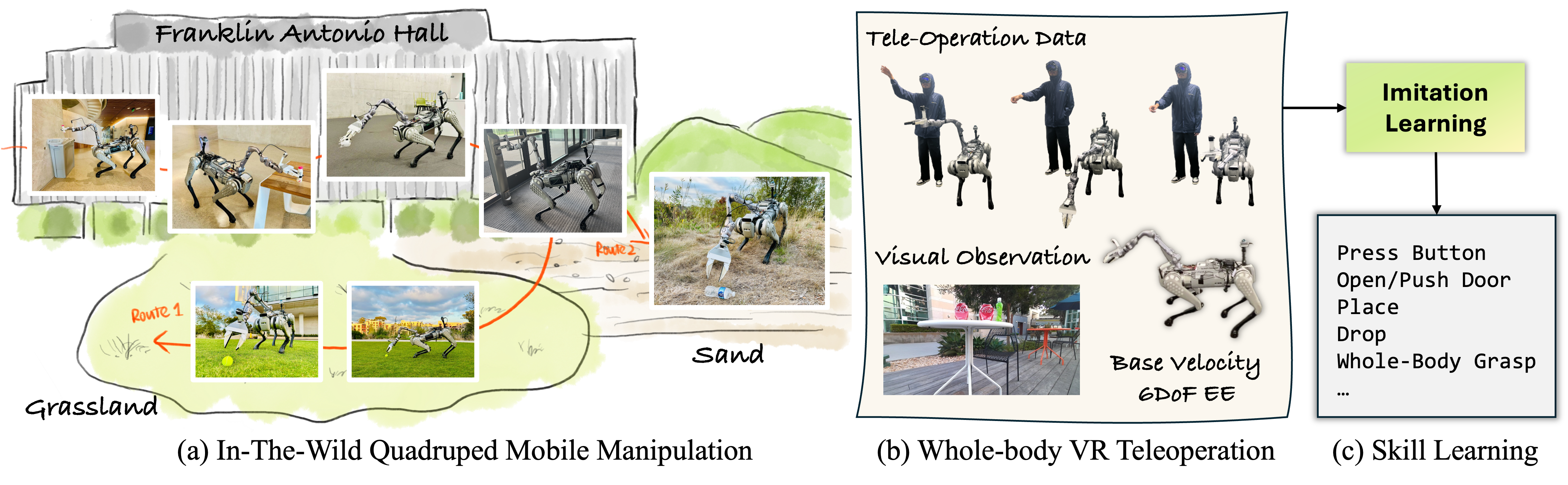}
    \caption{
    \acronym{} implements a framework for in-the-wild manipulation with a quadruped robot, which combines a whole-body controller and imitation learning for effective single-skill learning.
    (a) Long Horizon Loco-Manipulation in indoor as well as outdoor settings. (b) Teleoperation demonstration for collecting training data for imitation learning. (c) The constructed skill library with various skills, which can be composed by LLM planner to complete complex tasks.}
    \label{fig:teaser_fig}
\end{center}%
}]

\begin{abstract}

`In-the-wild' mobile manipulation aims to deploy robots in diverse real-world environments, which requires the robot to (1) have skills that generalize across object configurations; (2) be capable of long-horizon task execution in diverse environments; and (3) perform complex manipulation beyond pick-and-place. Quadruped robots with manipulators hold promise for extending the workspace and enabling robust locomotion, but existing results do not investigate such a capability. This paper proposes \textit{\acronym{}} with three components to address these issues: (1) adaptation of learned low-level controller for VR-enabled whole-body teleoperation and traversability; (2) \textit{\acronym{}-Skill} --- a library of generalizable visuomotor skills acquired via imitation learning or heuristics and (3) \textit{\acronym{}-Planner} --- an interface of learned skills that allow LLM planners to coordinate skills for long-horizon tasks. We demonstrate the importance of high-quality training data by achieving higher grasping success rate over existing RL baselines using only tens of demonstrations. \acronym{} exploits CLIP for language-conditioned imitation learning that empirically generalizes to objects unseen in training demonstrations. Besides extensive quantitative evaluation, we qualitatively demonstrate practical robot applications, such as cleaning up trash in university hallways or outdoor terrains, operating articulated objects, and rearranging items on a bookshelf.

\end{abstract}

\section{Introduction}

Practical robot mobile manipulation requires generalizable skills and long-horizon task execution. Consider a scenario where a mobile robot is deployed out-of-box at a family house. The robot is tasked with daily chores including collecting the trash around the house and grabbing something for human. To accomplish these tasks, the robot needs skills that generalize to unseen objects and a planner capable of compositing skills over a long horizon.

Existing methods~\cite{liu2024-vbc,liu2024-okrobot,qiu2024-geff,zhi2024-comerobot,fu2024mobilealoha,gu2024-conceptgraphs, xiong2024adaptive} have approached mobile manipulation from two primary directions. Modular methods~\cite{liu2024-okrobot, qiu2024-geff, zhi2024-comerobot} aim at designing decoupled perception-planning modules. With advances in large-scale vision models~\cite{kirillov2023segment, radford2021-CLIP, liu2023-groundingdino}, recent modular methods~\cite{qiu2024-geff,liu2024-okrobot} exhibit strong generalizability in perception to an open set of language-specified objects. However, their planning modules~\cite{chang2023goat,liu2024-okrobot,qiu2024-geff, gu2024-conceptgraphs} often rely on heuristic-based motion planning, limiting tasks to mostly simple pick-and-place. End-to-end approaches~\cite{liu2024-vbc,fu2024mobilealoha,zhao2023aloha,ha2024-umionlegs,chi2024-umi-gripper,chi2023-diffusionpolicy}, on the other hand, use learned policies to enable robot with complex actions. They, however, often hold a strong assumption of the small training-testing distribution gap ({\it e.g.,} sim2real~\cite{liu2024-vbc} or intra-class variation~\cite{fu2024mobilealoha}) and thus do not show strong generalizability. In addition, policies learned via imitation learning are prone to compounding error~\cite{zhao2023aloha,iyer2024-openteach} over long-horizon execution. Thus, these learned skills should be designed to be as atomic as possible for both generalizability and accuracy.

This paper investigates \textit{in-the-wild mobile manipulation} that addresses these issues for real-world deployment. Specifically, in-the-wild manipulation requires the robot to have skills that (1) generalize across texture, lighting, and diverse environments; (2) are capable of long-horizon execution; and (3) perform complex manipulation beyond pick-and-place. 

To this end, we propose \acronym{}. For generalizability, \acronym{} enables language-conditioned imitation learning (\acronym{}-Skill). Building upon ACT~\cite{zhao2023aloha, fu2024mobilealoha}, \acronym{}-Skill improves generalizability via pre-trained CLIP and composable skills. Instead of simply using CLIP features~\cite{chi2024-umi-gripper,ha2024-umionlegs}, we apply a reparameterization trick~\cite{zhou2022-MASKCLIP} to CLIP to compute probability maps given object text query as an auxiliary input. We then use VR teleoperation~\cite{cheng2024-opentv, ding2024-bunnyvisionpro} to collect human demonstrations to acquire complex skills such as non-prehensile manipulation. We adapt a learned low-level controller~\cite{liu2024-vbc} for VR-based whole-body teleoperation, which significantly increases the robot workspace and reduces the demonstration cost by 26.9\% compared to the decoupled strategy. Finally, based on a library of acquired generalizable and atomic skills, \acronym{} provides a language interface (\acronym{}-Planner) that allows interfacing with LLMs to composite skills for long-horizon execution.

In summary, our contributions are:

\begin{itemize}
    \item A generic framework with techniques that allow generalizable language-condition imitation learning (\textit{\acronym{}-Skill}) with interfacing to the LLM planner (\textit{\acronym{}-Planner}).
    \item Demonstrations of in-the-wild mobile manipulation tasks with full-stack and systematic deployment of the proposed framework.
    \item Comprehensive evaluation and ablation for the proposed technique, which paves the way for future study.
\end{itemize}

\section{Related Work}
\label{sec:related_works}

\begin{figure*}[t]
    \centering
    \includegraphics[width=.96\textwidth]{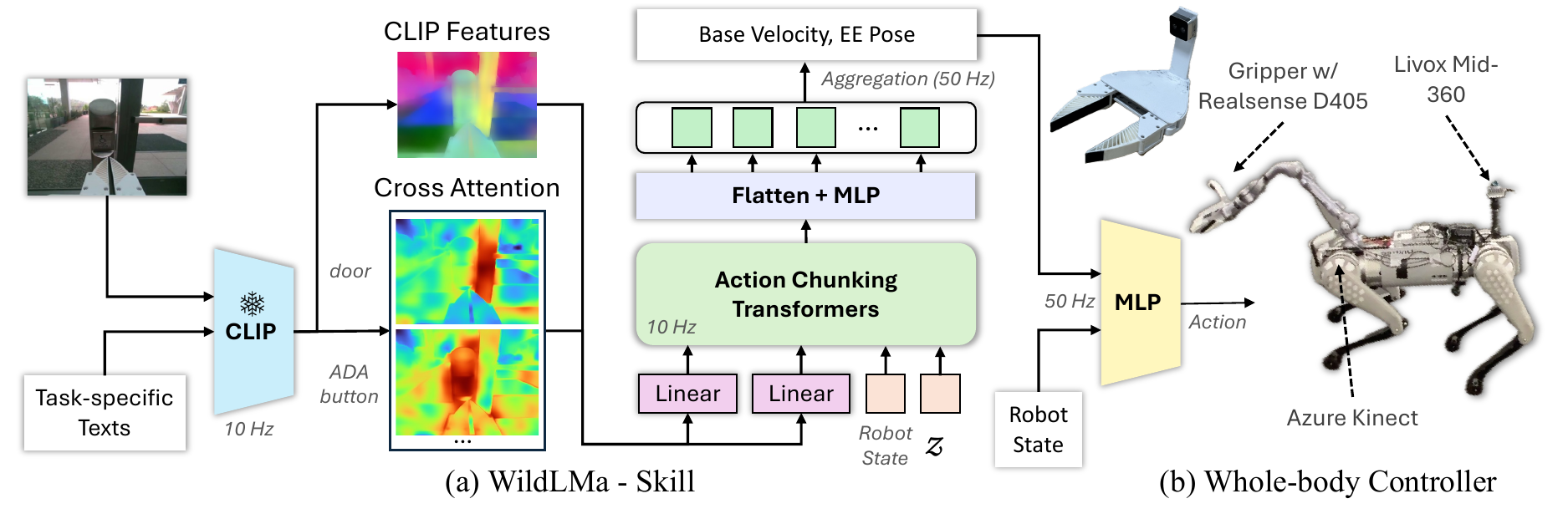}
    \vspace{-5pt}
    \caption{\textbf{Overview of \acronym{} models and robot setups.} (a) \acronym{} takes a frozen CLIP model to encode task-specific texts and visual observations; (b) Our robot platform is a Unitree B1 quadruped combined with a Unitree Z1 arm and a 3D-printed gripper, with two RGBD cameras and one lidar mounted on.
    }
    \vspace{-10 pt}
    \label{fig:clipact_wholebody_overview}
\end{figure*}

\paragraph{\textbf{Mobile Manipulation}}
Mobile manipulation has gained increasing attention for its vision of enabling robots to perform diverse practical tasks. In terms of hardware configurations, wheeled robots have made substantial strides~\cite{wu2023tidybot,xiong2024adaptive,sun2022fully,ahn2022can,lew2023robotic,liu2024-okrobot,zhi2024-comerobot} for its reliable base movement~\cite{hellorobot-website}, while recently, legged robots have also gained more interest for its robust locomotion~\cite{cheng2024-extremeparkour,yang2023-legged-prior} and the extended workspace via whole-body arm-base coordination~\cite{liu2024-vbc,ha2024-umionlegs,portela2024-compliant,fu2023deep}.

Categorized by methodology, existing methods can be divided into modular methods and end-to-end methods. Recent modular approaches~\cite{liu2024-okrobot,zhi2024-comerobot,qiu2024-geff,zhang2023gamma,yokoyama2023asc,arcari2023bayesian,zhang2024-adaptigraph,ji2024graspsplats,liu2024-dynamem} design decoupled perception-planning strategy. In particular, perception~\cite{liu2024-okrobot,qiu2024-geff,zhi2024-comerobot} are often done by applications of vision foundation models~\cite{gu2021open,kirillov2023segment,radford2021-CLIP,liu2023-groundingdino}; whereas grasping are done by off-the-shelf pose prediction models ({\it e.g.,} GraspNet~\cite{fang2020-graspnet}) and IK solver~\cite{rosmoveit}. Despite strong perception designs, modular methods are mostly limited to simple pick-and-place tasks. On the other hand, end-to-end approaches use Reinforcement Learning (RL)~\cite{gu2023multiskill,xia2021relmogen,xiong2024adaptive,liu2024-vbc,pan2024-roboduet} or Imitation Learning (IL)~\cite{fu2024mobilealoha,shafiullah2023bringing,ha2024-umionlegs} to enable complex tasks beyond pick-and-place such as articulated manipulation~\cite{xiong2024adaptive,bharadhwaj2024-track2act} and non-prehensile manipulation~\cite{fu2024mobilealoha,ha2024-umionlegs}. However, these work often fall short when training-testing distributions mismatch.

Most closely related to our work, \citet{yokoyama2023asc} proposed to use sim2real RL for in-the-wild mobile manipulation. However, they do not investigate manipulation tasks other than simple pick-and-place. \acronym{} uses imitation learning to learn diverse skills with generalizability, task complexity, and long-horizon run for in-the-wild execution.

\paragraph{\textbf{Long-horizon Mobile Manipulation}}
For robots to assist with real-world tasks such as cleaning up home, they need to be capable of dealing with long-horizon mobile manipulation, where independent skills are planned and triggered to complete given goals. Existing methods rely on sampling-based planning~\cite{siddharth2014-interfacedplanning,garrett2020pddlstream}, RL~\cite{yokoyama2023asc,yenamandra2023-homerobot,liang2024-skilldiffuser,gu2023multiskill}, and Large Language Models (LLMs)~\cite{rana2023-sayplan,song2023-llm-planner,huang2022-zeroshot-planner,gu2024-conceptgraphs} to coordinate skill primitives for long-horizon task execution. Recent work~\cite{rana2023-sayplan,song2023-llm-planner,huang2022-zeroshot-planner,gu2024-conceptgraphs} have found that LLM-based methods, especially Large-Mutlimodal Models (LMMs)~\cite{cheng2024spatialrgpt,openai2023gpt4}, are promising to serve as effective planners for embodied agents, where the research efforts are centered around hierarchical search~\cite{rana2023-sayplan} and re-planning~\cite{zhi2024-comerobot}. \acronym{} is intended to be orthogonal to these existing work in LLM planner. Instead of studying the planning capability, we investigate the potential of interfacing LMMs with skills acquired via imitation learning for practical applications. 

\paragraph{\textbf{Imitation Learning}}
Imitation learning has demonstrated promising results through learning from real-world expert demonstrations~\cite{fu2024mobilealoha,wong2022error,shafiullah2023bringing,chi2023-diffusionpolicy,ha2024-umionlegs,cheng2024-opentv,yang2024-ace,ding2024-bunnyvisionpro}.
Investigated for decades since the 80s~\cite{pomerleau1988alvinn}, behavior cloning~\cite{pomerleau1988alvinn,bojarski2016end} is one of the most commonly used imitation learning approach that learns an end-to-end mapping from observations to actions. Recently, researchers have shown that this classic approach not only allows complex manipulations~\cite{fu2024mobilealoha,cheng2024-opentv,ding2024-bunnyvisionpro,zhao2024-alohaunleashed}, but also holds the potential that scaling up training data with low-cost hardware~\cite{chi2024-umi-gripper,ha2024-umionlegs,yang2024-ace,zhao2023aloha,wu2023-gello,shafiullah2023bringing,fu2024mobilealoha} will lead to generalizable policies.
Similar to existing work~\cite{ding2024-bunnyvisionpro,cheng2024-opentv,shen2023-f3rm,qin2023-anyteleop,he2024omnih2o}, we also use VR devices to collect expert demonstrations that minimize the expert-agent observation gap~\cite{zhao2023aloha}. To reduce the cross-embodiment gap between the human operator and the quadruped robot, we combine the VR demonstration with learned whole-body controller. In addition, we also improve the vanilla ACT model~\cite{zhao2023aloha,fu2024mobilealoha} to support language-conditioned imitation learning that is more generalizable with autonomous termination.

\paragraph{\textbf{Whole-body Control}}
Quadruped Whole-Body Control (WBC) draws inspiration from the natural motions of animals to extend the robot workspace via arm-base coordination. The WBC capability is usually achieved via model-based hierarchical trajectory optimization~\cite{bellicoso2019alma, ma2022combining, sleiman2023versatile, zimmermann2021go, zhou2022teleman} or sim2real RL~\cite{fu2023deep,liu2024-vbc,ha2024-umionlegs}.
Our work is based on the low-level controller proposed by VBC~\cite{liu2024-vbc}, which designed a bi-level RL paradigm with a low-level whole-body controller.
Notably, some existing work has also attempted to combine teleoperation with whole-body control for quadruped robots~\cite{zhou2022teleman,portela2024-compliant} but does not investigate learning skills from teleoperation.
Most related to our work, \citet{ha2024-umionlegs} demonstrated whole-body imitation learning with handheld data collection hardware. The main differences between our work and \citet{ha2024-umionlegs} are (1) the data collected without teleoperating the robot can include only wrist camera observation, which may lead to worse performance than multi-camera setup as we empirically verify and (2) \citet{ha2024-umionlegs} focus on execution of short tasks; while we investigate in-the-wild mobile manipulation with long horizon task execution.

\vspace{-5pt}
\section{Method}

\acronym{} designs three components to address challenges for in-the-wild mobile manipulation. Sec.~\ref{sec:whole_body_teleop} describes adapting a whole-body controller to support efficient teleoperation and more diverse real-world tasks. In Sec.~\ref{sec:wildact_skill}, we propose \textit{\acronym{}-Skill}, which modifies the pre-trained CLIP model~\cite{radford2021-CLIP} for generalizable imitation learning. \acronym{}-Skill then constructs a skill library consisting of learned skills and analytical skills ({\it e.g.,} navigation). Finally, \textit{\acronym{}-Planner} (Sec.~\ref{sec:wildact_planner}) interfaces \acronym{}-Skill with an LLM planner to carry out long-horizon execution.

\subsection{Whole-body VR Teleoperation}
\label{sec:whole_body_teleop}

Recent imitation learning methods have benefited from improved data collection methods via VR/AR-based teleoperation~\cite{cheng2024-opentv,ding2024-bunnyvisionpro,qin2023-anyteleop,he2024omnih2o}. However, though human operators can naturally tele-operate bipedal humanoid robots~\cite{he2024omnih2o,cheng2024-opentv,fu2024-humanplus}, it is non-trivial to tele-operate quadruped robots due to the embodiment gap~\cite{padalkar2023-open-X} between two-legged human and quadruped robots inspired by four-legged animals.

To reduce the need for the tele-operator to consider both the base movement and the arm movement, we propose to use a whole-body controller~\cite{liu2024-vbc} that allows smooth arm-base coordination for the robot. In particular, we use the low-level whole-body policy developed by \citet{liu2024-vbc}. Trained with RL, the learned whole-body controller takes in \textit{base commands} (linear velocity and angular velocity) and \textit{6DOF end effector pose} w.r.t. the arm base. The policy outputs arm and base joint commands for coordinated movement that extends the workspace (illustrated in Fig.~\ref{fig:teaser_fig}).

Based on the pre-trained low-level controller, we then design an interface for human users to teleoperate the robot. We use the OpenTV framework~\cite{cheng2024-opentv} with Apple Vision Pro, which allows real-time video streaming, tracking of 6DOF poses of head and hands, and 3D gesture keypoints. To minimize the expert-agent observation gap~\cite{zhao2023aloha}, the tele-operator gets real-time streams of the robot's head camera views and wrist camera views.

\begin{figure*}[t]
    \centering
    \includegraphics[width=.96\textwidth]{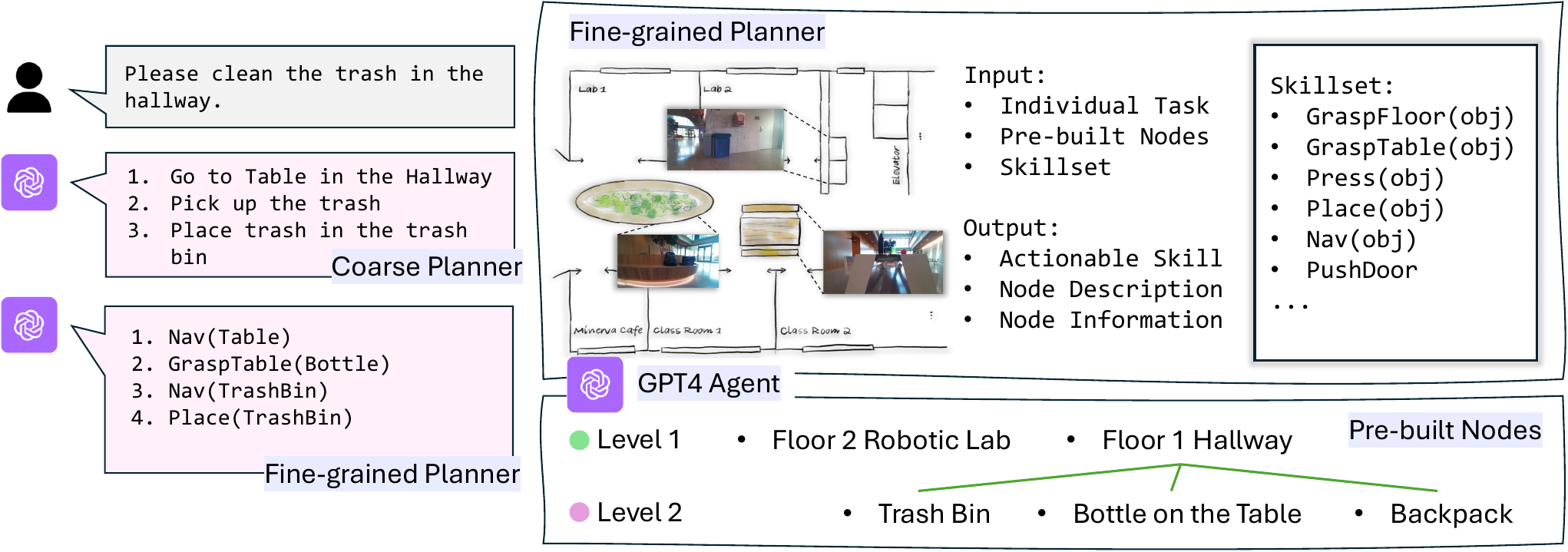}
    \caption{Overview of \acronym{}-planner. Given a constructed hierarchical scene graph, \acronym{}-planner adopts a coarse-to-fine searching mechanism to determine node traversal and structured actions to take.}
    \vspace{-10pt}
    \label{fig:WildACT_planner_overview}
\end{figure*}

To translate tele-operator movement to robot movement, we linearly transform the operator's right wrist pose (relative to their initial hand pose) $T_{right} \in \mathrm{SE}(3)$ into the relative end effector pose $T_{ee} \in \mathrm{SE}(3)$. We scale the translations with a constant $s_c$, as we find that the workspace of the Z1 arm is slightly larger than average human arms. More concretely, let $\mathbf{R}_{right}$ be the rotational component and $\mathbf{t}_{right}$ be the translational component of $T_{right}$, $T_{ee}$ is given by,%
\begin{equation}%
    T_{ee} = \begin{bmatrix}
        \mathbf{R}_{right} & s_{c} \cdot \mathbf{t}_{right} \\
        \mathbf{0}^{\intercal} & 1
    \end{bmatrix}\,.
\end{equation}%
The gripper open-close actions are then naturally mapped from the pinching of the thumb and the index finger (via 3D keypoints). The whole-body controller automatically controls the base rotation to coordinate with the arm. In turn, the tele-oeprator's left wrist governs planar base movements ({\it e.g.,} angular and linear velocities). When the tele-operator pinches their left fingers, VR tracks the pose $T_{left}$ as a virtual joystick with deadzone ($x_{th} = 5cm$). We find this simple base command mapping sufficient for the tasks involved.

\subsection{\acronym{}-Skill}
\label{sec:wildact_skill}

\textit{\acronym{}-Skill} contains skills from two categories: skills acquired via imitation learning and with analytical planners.

\paragraph{\acronym{}-Skill - Imitation Learning}

The collected real-world demonstrations can be turned into autonomous skills with existing behavior cloning methods~\cite{fu2024mobilealoha,cheng2024-opentv}. Many existing methods, however, struggle to generalize to novel environments~\cite{fu2024mobilealoha}. To improve the generalizability of learned skills without expensive demonstration collection cost, we propose to adapt pre-trained CLIP~\cite{radford2021-CLIP} to ACT~\cite{zhao2023aloha} for imitation learning of individual skills.

\noindent{\textbf{Improving Generalizability with CLIP.}} We encode camera observations with a frozen CLIP visual backbone. Instead of simply using intermediate CLIP features as in~\cite{chi2024-umi-gripper,ha2024-umionlegs}, we apply MaskCLIP~\cite{zhou2022-MASKCLIP}, a reparameterization trick to generate image-text cross attention map. More concretely, let $\Omega$ be the space of RGB images. The original CLIP~\cite{radford2021-CLIP} is a mapping function $f_{visual}(\cdot): \Omega \mapsto \mathbb{R}^{C}$, where $\mathbb{R}^{C}$ is the image-text embedding space learned from contrastive learning~\cite{radford2021-CLIP}. MaskCLIP modifies the network architecture to a new mapping function $g_{visual}(\cdot): \Omega \mapsto \mathbf{H} \times \mathbf{W} \times \mathbb{R}^{C}$, which is a feature map aligned to the CLIP embedding space $\mathbb{R}^{C}$ (illustrated in Fig.~\ref{fig:clipact_wholebody_overview}).

\begin{table*}[t]
\centering
 \begin{tabular}{l c  c  c  c  c  c  c}
 \toprule
 & \multicolumn{2}{c}{Tabletop Grasping} & \multicolumn{2}{c}{Button Pressing} & \multicolumn{2}{c}{Ground Grasping}
 \\
 \cmidrule(lr){2-3} \cmidrule(lr){4-5} \cmidrule(lr){6-7}
 Method & I.D. & O.O.D. & I.D. & O.O.D & I.D. & O.O.D & Avg. Succ.\\
 \midrule
 \acronym{} (Ours) & \textbf{94.4\%} & 75\% & \textbf{80\%}
    & \textbf{57.5\%} & \textbf{60\%} & \textbf{60\%} & \textbf{71.2\%}
 \\
 ACT (Mobile ALOHA)~\cite{fu2024mobilealoha} & 77.8\% & 19.4\% & 55\% & 25\% & 60\% & 30\% & 40.8\%
 \\
 OpenTV~\cite{cheng2024-opentv} & 88.9\%
    & \textbf{77.8\%} & 75\% & 25\% & 50\% & 50\%& 64.4\%
 \\
 VBC~\cite{liu2024-vbc} & 50\%\textcolor{red}{$^{*}$}
    & 50\%\textcolor{red}{$^{*}$} & NA\textcolor{red}{$^{\dagger}$} & NA\textcolor{red}{$^{\dagger}$} & 43.8\%\textcolor{red}{$^{*}$} & 43.8\%\textcolor{red}{$^{*}$} & 46.9\%
 \\
 GeFF~\cite{qiu2024-geff} & 55.6\%\textcolor{red}{$^{*}$}
    & 55.6\%\textcolor{red}{$^{*}$} & NA\textcolor{red}{$^{\dagger}$} & NA\textcolor{red}{$^{\dagger}$} & NA\textcolor{red}{$^{\dagger}$} & NA\textcolor{red}{$^{\dagger}$} & 55.6\%
 \\
 \bottomrule
 \end{tabular}
 \caption{\textbf{Success rate of autonomous skill execution}. Imitation learning methods outperform RL~\cite{liu2024-vbc} and zero-shot method~\cite{qiu2024-geff} on comparable tasks. Both OpenTV and \textit{\acronym{}} achieve noticeably higher success rates in the challenging O.O.D. setting. \textcolor{red}{$^{\dagger}$}: methods involve learned/manual policies that are not trivially applicable to the task settings. \textcolor{red}{$^{*}$}: Method does not differentiate object sets and success rates are averaged on I.D. and O.O.D. object sets.} 
 \vspace{-1em}
 \label{table:main_results}
\end{table*}

\noindent{\textbf{Image-text Cross Attention.}} Consistent with findings by \citet{chi2024-umi-gripper}, we found that the adaptation of CLIP~\cite{radford2021-CLIP} improves the performance. However, when tested with objects unseen in the training demonstrations, the success rate is still unsatisfactory. Thus, we propose to make the acquired skills \textit{language-conditioned} by introducing cross-attention. We provide task-specific texts during both the training and testing time ({\it e.g.,} for the ADA door button-pressing task, we use `door' and `ADA button') with CLIP text embedding $f_{text}(\cdot): \textit{Text} \mapsto \mathbb{R}^{C}$. With slight abuse of notation, the text vector can then be compared with the CLIP feature map via cosine similarity%
\begin{equation}%
    \textsc{CrossAtt}(\cdot,\cdot) = \frac{g_{visual}(\cdot) f_{text}(\cdot)}{\left| \left| g_{visual}(\cdot) \right| \right| \left| \left| f_{text}(\cdot) \right| \right|},
\end{equation}%
where the comparison is done independently on the pixel level. The resulting similarity is comparable to the probability map of text queries. We apply dropout~\cite{srivastava2014-dropout} to cross-attention during training to avoid over-reliance on attention.

\noindent{\textbf{Autonomous Termination.}} To autonomously terminate skills, in order to hand control back to high-level planners, we add a virtual \textit{`end'} action signal prediction. Empirically, adding the end signal to only the end of the episode does not work, as the supervision is too sparse. Our proposed solution is to implement a buffer of end signal for every skill such that the last $n=10$ frames of the demonstrations carry the end signal. During deployment, we use a sliding window detector to terminate task execution if the end signal is greater than $\tau = 0.8$ for 10 consecutive predictions.

\paragraph{\acronym{}-Skill - Analytical Planning}

In this paper, we learn all manipulation-related skills with imitation learning. For the base-only skill ({\it i.e.,} navigating from a known location to another known location), we implement it with analytical planning.

\subsection{\acronym{}-Planner}
\label{sec:wildact_planner}

The \textit{\acronym{}-Skill} module provides skills that can be composed for long-horizon execution, which is intentionally designed to be agnostic of the high-level planner. Here, we propose \textit{\acronym{}-Planner}, a simple LLM-based planner to show how learned skills can be composed.

\noindent{\textbf{Initial Mapping.}} We implement a LiDAR-based SLAM system using FAST-LIO~\cite{xu2021-fastlio} and DLO~\cite{chen2022-DLO} to obtain consistent robot pose estimation in the world frame. We manually annotate pose-level waypoints ({\it e.g.,} stand in front of receptacles) and connectivity for task execution. The robot stands at every waypoint to capture images with its head camera. To automatically annotate the semantics of each waypoint, GPT4-V~\cite{openai2023gpt4} provides high-level descriptions of images and lists of objects of each waypoint. We manually create abstract nodes ({\it e.g.,} a room with multiple pose-level waypoints) to construct a hierarchical graph for searching. Note that off-the-shelf scene graph construction methods~\cite{maggio2024-clio,hughes2022-hydra,gu2024-conceptgraphs} can potentially replace this step.

\noindent{\textbf{Hierarchical Long-horizon Planning.}} We adopt a hierarchical coarse-to-fine approach to translate template-free commands into detailed, actionable robot skills.

\textit{Coarse Planner.} Using Chain of Thought~\cite{wei2022-CoT}, the coarse planner receives template-free instructions and decomposes them into individual tasks. For instance, the command `clean the trash in the hallway' can be decomposed into tasks `navigate to hallway', `pick up the trash', `navigate to trash bin', and `place trash in the trash bin'. 

\begin{table}[t]
\centering
 \begin{tabular}{l c c c}
 \toprule
 Pipeline & Collect \& Drop Trash & Shelf Rearrangement\\
 \midrule
 \acronym{} (Ours)
    & \textbf{7/10} & \textbf{3/10}
 \\
 ACT~\cite{fu2024mobilealoha,zhao2023aloha}
    & 0/10 & 0/10
 \\
 \bottomrule
 \end{tabular}
 \caption{\textbf{Evaluation of long-horizon execution}. Given a few training demonstrations (10), \acronym{} improves long-horizon task success rate via (1) improved generalizability of single skill and (2) divide-and-conquer.}
 \vspace{-1em}
 \label{table:long_horizon_evaluation}
\end{table}

\textit{Fine-grained Planner.} The fine-grained planner invokes actionable skills at particular nodes given individual tasks generated by the coarse planner. The fine-grained planner has prior knowledge of the robot’s skill library (shown in Fig.~\ref{fig:WildACT_planner_overview}) and nodes constructed in the initial mapping stage. For each task, the agent uses a breadth-first search (BFS) approach to search nodes and identify the optimal goal node. During this stage the LLM acts as a heuristic evaluator, estimating the likelihood of a node being the most likely location related to the task, based on the semantic context and objects present at the node. Once the target node is identified, the planner constructs a plan detailing the navigation and manipulation sequence drawn from the pre-defined skill library.

\section{Experiments}

{\bf Hardware Platforms.} We use the Unitree B1 quadruped robot with a Unitree Z1 arm. We replace the beak-like default Z1 end effector with a 3D-printed parallel soft gripper, which was adapted from UMI-Gripper~\cite{chi2024-umi-gripper} to directly operate the gripper with gear rotations. For perception, an Azure Kinect camera is mounted on the robot's head, and an Intel Realsense D405 is used as the in-wrist camera. A LIVOX MID-360 LiDAR is installed at the robot's tail for enhanced localization during navigation.

\textbf{Implementation Details.} 
The \acronym{}-Skill module independently trains weights for each skill (with 30-60 demonstrations each acquired via tele-operation). The head/wrist RGB observations are processed through a CLIP~\cite{radford2021-CLIP} ViT-B/16 encoder with MaskCLIP~\cite{zhou2022-MASKCLIP} re-parameterization. Task-specific texts are then compared with the feature maps to generate cross-attention, where texts may differ in training sequences and testing run. For navigation between given waypoints, we implement a PD-based waypoint follower.
\acronym{}-Planner requires geometric annotations of nodes and edges. For efficiency, the spatial locations of nodes are annotated by operating the robot to turn 360 degrees during the initial scene scanning, and the edges are made between physically adjacent nodes with no obstacle in between.





\begin{table}[t]
\centering
 \begin{tabular}{l c  c  c}
 \toprule
 Backbone & In Dist. & Out of Dist. & Avg. Succ.\\
 \midrule
 CLIP~\cite{radford2021-CLIP}
    & 83.3\%& 69.4\%& 76.4\%
 \\
 ResNet~\cite{zhao2023aloha}\textcolor{red}{$^{\star}$}
    & 77.8\% & 19.4\% & 48.6\%
 \\
 DinoV2~\cite{oquab2023-dinov2}
    & \textbf{88.9\%} & \textbf{77.8\%} & \textbf{83.3\%}
 \\
 \bottomrule
 \end{tabular}
 \caption{\textbf{Ablation of different visual encoders} pre-trained with different objectives. The evaluation is done on the object-grasping tasks. \textcolor{red}{$^{\star}$}: we followed ACT~\cite{fu2024mobilealoha,zhao2023aloha} to use ImageNet-pretrained ResNet-18 as the encoder, which has fewer parameters.}
 \vspace{-1.5em}
 \label{table:backbone_ablation}
\end{table}

\begin{table*}[t]
\centering
\small 
\setlength{\tabcolsep}{4pt} 
\begin{tabular}{l c c | c c | c c}
    \toprule
    & \multicolumn{2}{c|}{Whole-body (Ours)} & \multicolumn{2}{c|}{Decoupled Control} & \multicolumn{2}{c}{W/o Whole-body (Arm Only)} \\
    \cmidrule(lr){2-3} \cmidrule(lr){4-5} \cmidrule(lr){6-7}
    Metric & Ground Grasping & Rearrange Shelf & Ground Grasping & Rearrange Shelf & Ground Grasping & Rearrange Shelf \\
    \midrule
    Average Time  & 21.87s & 27.25s & 37.35s  & 29.81s & - & 27.88s \\
    Success Rate  & 95\% & 70\% & 80\% & 40\% & 0\% & 70\% \\
    \bottomrule
\end{tabular}
\caption{\textbf{Comparison of success rate and completion time} for our whole-body controller, decoupled control with manual base pitching and arm control implemented via Unitree SDK, and arm-only policies. Four teleoperators are tasked to manipulate objects at various heights for three trials in each task.}
\vspace{-10px}
\label{table:wholebody_efficiency_combined}
\end{table*}

{\bf Experimental Protocol.} We define two experiment settings to investigate the generalizability of skills learned via imitation learning~\cite{cheng2024-opentv,fu2024mobilealoha,ding2024-bunnyvisionpro}. The \textit{in-distribution (I.D.)} setting tests the learned skills with backgrounds and object arrangements approximately similar to the training demonstrations. Note that, to make the setting more realistic, we do not enforce identical robot positioning and lighting conditions even in I.D. settings. The \textit{Out-of-distribution (O.O.D.)} setting permutes the testing objects (placement/texture), receptacles, and background environments for learned skills. 

\begin{table}[t]
\centering
 \begin{tabular}{l c c c}
 \toprule
 Camera & Tabletop Grasping & Button Pressing & Door Opening\\
 \midrule
 Head + Wrist
    & \textbf{94.4\%} & 80\% & \textbf{70}\%
 \\
 Head Only
    & 27.8\% & 75\% & 30\%
 \\
 Wrist Only
    & 83.3\% & \textbf{85\%} & 10\%
\\
 \bottomrule
 \end{tabular}
 \caption{\textbf{Ablation of input visual modality}. Tasks that involve occlusion significantly benefit from multi-view setup.}
 \label{table:long_horizon_evaluation}
\end{table}

{\bf Baseline Implementation.} Besides ablating design choices of our components, we implement several baselines to validate the efficacy of \acronym{}. To compare with existing imitation learning methods, we choose Mobile ALOHA~\cite{fu2024mobilealoha} which uses ACT~\cite{zhao2023aloha} with ResNet-18 and OpenTV~\cite{cheng2024-opentv} using ACT and DinoV2~\cite{oquab2023-dinov2}. Unless specifically noticed, these baselines use the same training data as \acronym{}. In addition, we compare two recent works in quadruped loco-manipulation~\cite{liu2024-vbc,qiu2024-geff} to compare \acronym{} against RL-based and zero-shot grasping methods. Note that both VBC~\cite{liu2024-vbc} and GeFF~\cite{qiu2024-geff} were designed for grasping, so they are not trivially applicable to non-prehensile manipulation such as button pressing.

\subsection{Evaluation}

We address important research questions in our evaluation:

\begin{itemize}

\item What advantages does \textit{\acronym{}-Skill} have compared to existing baselines in quadruped manipulation? [A1, A2]

\item How does \textit{\acronym{}-Planner} perform in long-horizon execution? [A3]

\item Are the design choices ({\it e.g.,} visual backbone and cross-attention) effective? [A4, A5]

\item Does whole-body control improve teleoperation? [A6]

\item What are the real-world applications of \acronym{}? [A7]

\end{itemize}

\noindent{\textbf{A1. \acronym{} outperforms recent imitation learning baselines.}} From Tab.~\ref{table:main_results}, we can see that \acronym{} achieves best overall success rate. Compared to vanilla ACT~\cite{zhao2023aloha,fu2024mobilealoha}, \acronym{} achieves slightly better success rates on I.D. setting and significantly better success rate on the O.O.D. setting. We reason this is because ResNet is vulnerable to changes in lighting and texture. OpenTV~\cite{cheng2024-opentv}, on the other hand, shows more robustness to these adversarial conditions due to its use of the recent DinoV2 backbone~\cite{oquab2023-dinov2}, but slightly underperforms our method.

\noindent{\textbf{A2. \acronym{} outperforms RL and zero-shot baselines.}} Due to less reliance on real-world demonstrations, RL and zero-shot baselines demonstrate less performance gap between I.D. and the O.O.D. settings in Tab.~\ref{table:main_results}. 
As an RL-based method, VBC~\cite{liu2024-vbc} suffers from sim2real gaps such as inaccurate contact modeling and cumulative sensor latencies. Therefore, VBC performs worse in real-world settings than its simulation counterpart. On the other hand, zero-shot modular methods such as GeFF~\cite{qiu2024-geff} do not naturally exhibit corrective behavior like learning-based methods, which are vulnerable to errors compounding from different modules.

\noindent{\textbf{A3. \acronym{} is capable of handling long-horizon manipulation under perturbations.}} Tab.~\ref{table:long_horizon_evaluation} validates the efficacy of \acronym{} in handling long-horizon tasks under certain perturbations. We include videos on the website. Our experiments include 20 training sequences with variations in robot positioning, lighting, and object placement in both the training and testing time. ACT fails entirely for long-horizon tasks when trained directly on a few sequences of demonstrations. On the other hand, \acronym{} successfully learns generalizable skills from a limited number of demonstrations to achieve better success rates for long-horizon execution.

\begin{table}[t]
\centering
 \begin{tabular}{l c c c}
 \toprule
 Backbone & In Dist. & Out of Dist. & Avg. Succ.\\
 \midrule
 w/ cross-attention (Ours)
    & \textbf{94.4\%} & \textbf{75\%} & \textbf{84.7\%}
 \\
 w/o cross-attention
    & 83.3\% & 69.4\% & 76.4\%
 \\
 \bottomrule
 \end{tabular}
 \caption{\textbf{Ablation of cross-attention} on the object-grasping tasks. Cross-attention improves both I.D. and O.O.D. setting by using additional task-specific information.}
 \vspace{-1em}
 \label{table:attention_mask_ablation}
\end{table}

\begin{figure}[t]
    \centering
    \includegraphics[width=.96\linewidth]{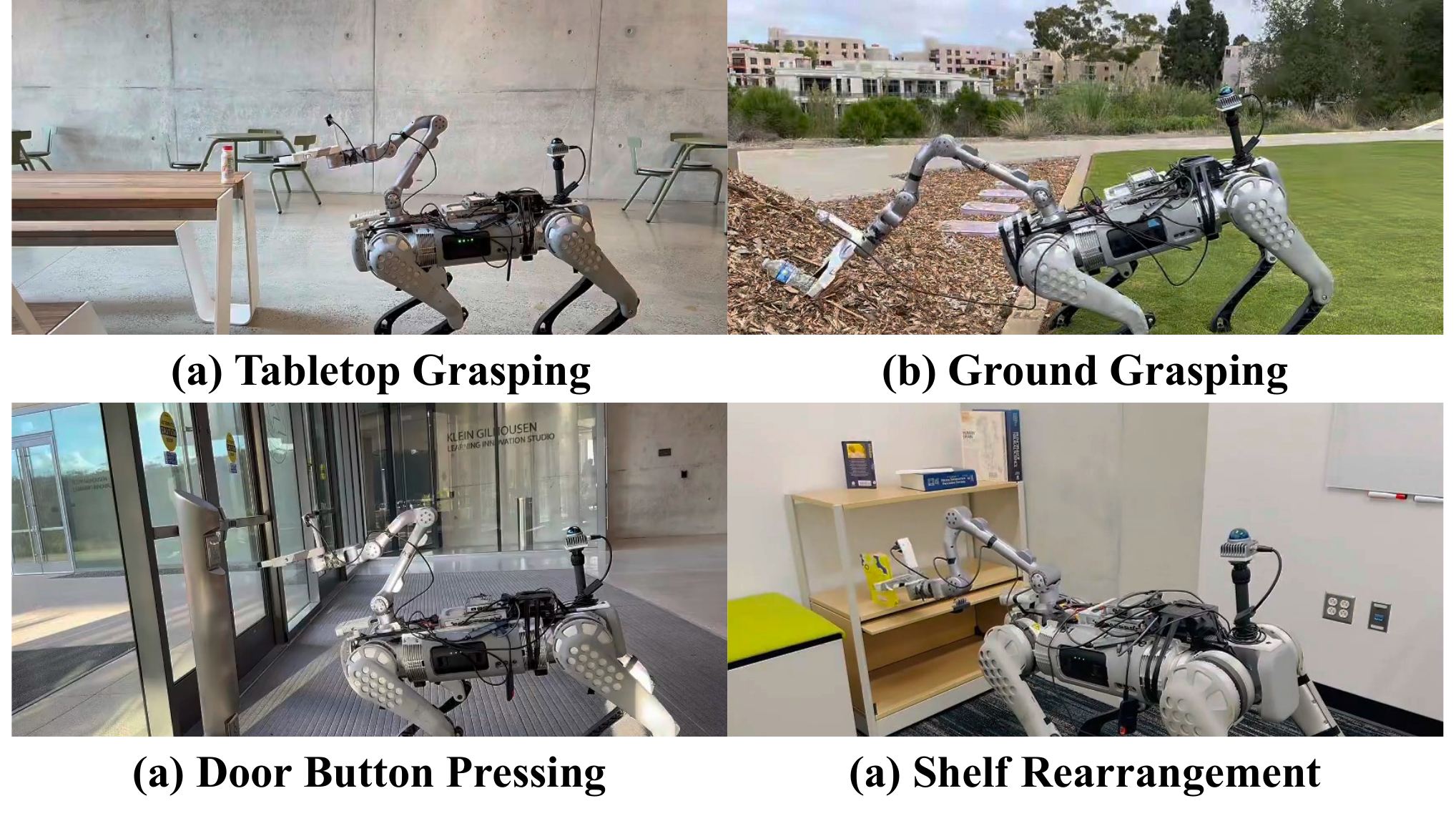}
    \caption{Qualitative illustrations of some evaluated tasks.}
    \vspace{-10pt}
    \label{fig:qual_fig}
\end{figure}

\noindent{\textbf{A4. Pre-trained Visual Backbones improve skill generalizability.}} We ablate the choice of visual backbones in Tab.~\ref{table:backbone_ablation}. CLIP~\cite{radford2021-CLIP} is the simple application of CLIP features without cross-attentions. While different backbones perform similarly in the I.D. setting, we see that frozen large models~\cite{radford2021-CLIP,oquab2023-dinov2} perform much better in the O.O.D. setting.

\noindent{\textbf{A5. Cross-attention significantly improves O.O.D. imitation learning.}} Tab.~\ref{table:attention_mask_ablation} shows the proposed cross-attention improves both the I.D. and O.O.D. performance of CLIP~\cite{radford2021-CLIP} module by introducing additional task-specific text prompts.

\noindent{\textbf{A6. Whole-body controllers enable efficient VR teleoperation of quadruped robots.}} The motivation for combining whole-body control and teleoperation is to improve teleoperation efficiency. To validate this point, we report the statistics of teleoperation in Tab.~\ref{table:wholebody_efficiency_combined}, which shows our learning-based controller outperforms the decoupled analytical controller from Unitree SDK. Since these tasks require reaching objects at various heights (toys and books at different levels of storage), teleoperation without whole-body control fails to grasp from the ground due to the limited workspace.

\noindent{\textbf{A7. \acronym{} allows the robot to learn diverse tasks.}} Besides Fig.~\ref{fig:qual_fig}, we provide more video demonstrations of \acronym{} in the supplementary video and the website.

\section{Conclusion}

In this paper, we present \acronym{}, a modular framework that includes (1) \textit{\acronym{}-Skill}, which implements a library of generalizable visuomotor skills that improve ACT~\cite{zhao2023aloha} for learning generalizable imitation learning skills; and (2) \textit{\acronym{}-Planner}, an interface that enables interactions between imitation learning skills and LLM planner to support long-horizon task execution. Furthermore, we deploy this framework on a quadruped robot controlled by a whole-body controller, which allows us to efficiently collect demonstration data and support extended workspace for diverse tasks. In summary, \acronym{} implements practical, generalizable skills, and long-horizon manipulation, which we hope will motivate future research toward in-the-wild mobile manipulation that facilitates real-world deployment of robots.

\section{Acknowledgement}

This project was supported, in part, by gifts from Amazon, Qualcomm and Meta.

\clearpage
\printbibliography

\end{document}